\documentclass[10pt,twocolumn,letterpaper]{article}

\usepackage[pagenumbers]{iccv} 

\usepackage{array}
\usepackage{multirow}
\usepackage{marvosym}
\PassOptionsToPackage{normalem}{ulem}
\usepackage{ulem}
\usepackage{amsmath}

\definecolor{iccvblue}{rgb}{0.21,0.49,0.74}
\usepackage[pagebackref,breaklinks,colorlinks,allcolors=iccvblue]{hyperref}

\def\paperID{7483}
\def\confName{ICCV}
\def\confYear{2025}

\title{Online Dense Point Tracking with Streaming Memory}

\author{Qiaole Dong and Yanwei Fu\textsuperscript{\Letter}\\
School of Data Science, Fudan University\\
{\tt\small \{qldong18, yanweifu\}@fudan.edu.cn}
}

\begin{document}
\maketitle
\begin{abstract}
Dense point tracking is a challenging task requiring the continuous tracking of every point in the initial frame throughout a substantial portion of a video, even in the presence of occlusions. Traditional methods use optical flow models to directly estimate long-range motion, but they often suffer from appearance drifting without considering temporal consistency. Recent point tracking algorithms usually depend on sliding windows for indirect information propagation from the first frame to the current one, which is slow and less effective for long-range tracking.
To account for temporal consistency and enable efficient information propagation, we present a lightweight and fast model with \textbf{S}treaming memory for dense \textbf{PO}int \textbf{T}racking and online video processing. The \textbf{SPOT} framework features three core components: a customized memory reading module for feature enhancement, a sensory memory for short-term motion dynamics modeling, and a visibility-guided splatting module for accurate information propagation. This combination enables SPOT to perform dense point tracking with state-of-the-art accuracy on the CVO benchmark, as well as comparable or superior performance to offline models on sparse tracking benchmarks such as TAP-Vid and RoboTAP. Notably, SPOT with 10$\times$ smaller parameter numbers operates at least 2$\times$ faster than previous state-of-the-art models while maintaining the best performance on CVO. We will release the models and codes at: \url{https://dqiaole.github.io/SPOT/}.
\end{abstract}

\section{Introduction}
\label{sec:intro}

A fine-grained and long-term analysis of motion, like what we experience in human vision, has been a long-standing goal in computer vision. This problem has typically been approached through optical flow~\cite{sun2018pwc, teed2020raft, jiang2021learning, dong2023rethinking} and point tracking~\cite{doersch2022tap, harley2022particle, karaev2023cotracker, doersch2023tapir}. While each type of motion facilitates various real-world applications~\cite{gao2020flow, sun2018optical, geng2022comparing, chen2024leap, vecerik2024robotap}, neither can fully capture dense, long-range motion in real time and on a frame-by-frame basis for streaming motion perception. Here, we consider dense point tracking, which is essentially the same task as long-range optical flow over many frames, and will use these terms interchangeably.

\begin{figure}
\centering
\includegraphics[width=\linewidth]{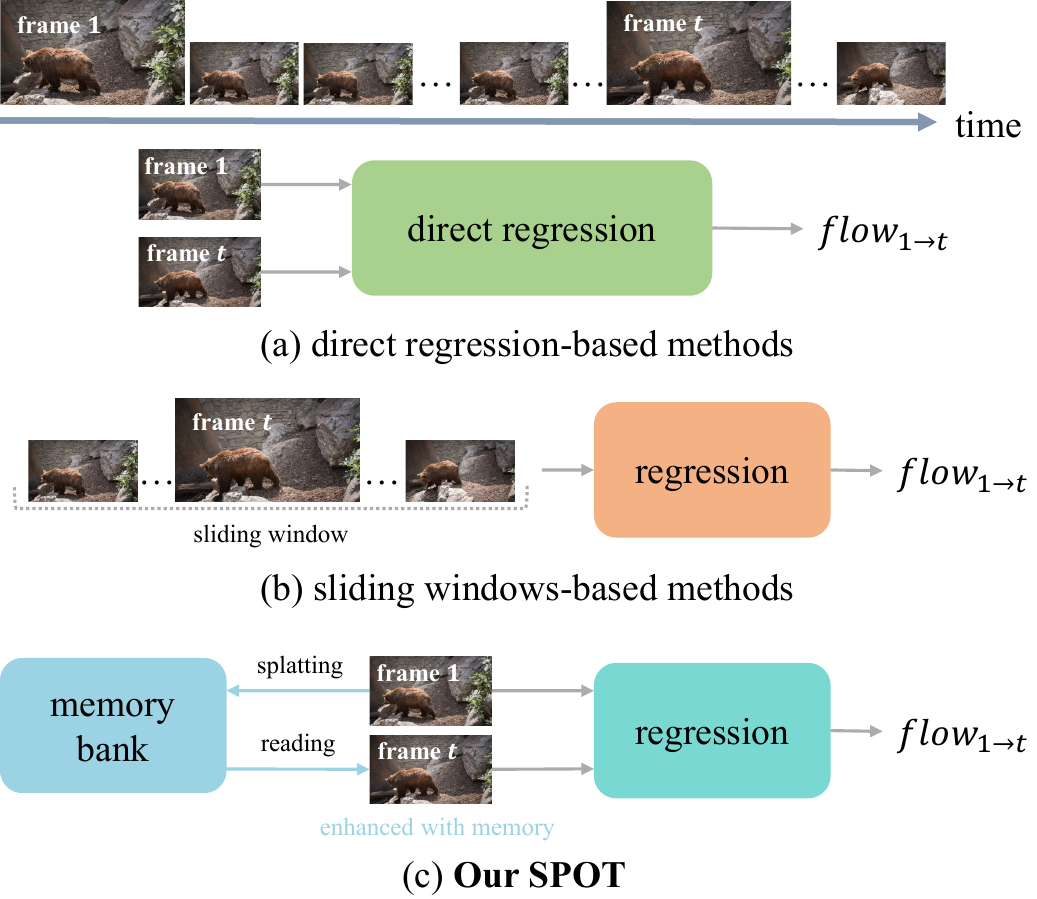}
\vspace{-0.3in}
\caption{Main frameworks for dense and long-range correspondence estimation. (a) Classical methods employ optical flow models for direct regression of long-range optical flow. (b) Recent sparse point tracking methods rely on sliding windows for information propagation and require offline processing. (c) Our SPOT introduces a fast streaming memory that stores information of first frame and enhances the current frame for better regression.\label{fig:teaser}}
\vspace{-0.2in}
\end{figure}

Classical methods typically utilize the optical flow~\cite{sun2018pwc, teed2020raft, jiang2021learning, dong2023rethinking} models, originally trained on adjacent frames, for direct regression of long-range flow as depicted in \cref{fig:teaser}. Recent work like TAP-Net~\cite{doersch2022tap} and MFT~\cite{neoral2024mft} has bridged the gap by training on long-range frame pairs and chaining reliable short-range flows respectively. However, they still suffer from appearance drifting and struggle with occlusions without considering the temporal consistency.

As the appearance information of physical points within the first frame needs to be propagated and compared to the one in the target frame, recent point tracking~\cite{harley2022particle, karaev2023cotracker, xiao2024spatialtracker} methods typically employ sliding windows with offline processing, \ie, necessitating future frames for improving the accuracy. Despite the super performance of these models, they require a huge amount of time to track every pixel. Implicit information propagation through sliding windows is also inefficient and less effective for long-range tracking. Recent hybrid method DOT~\cite{le2024dense} combines optical flow and sparse point tracking for fast and dense point tracking. However, it still inherits the disadvantage of base methods. This raises a question: \emph{Is it possible to achieve both high accuracy and efficiency based on past observation solely?}

We present a novel framework with \textbf{S}treaming memory for online dense \textbf{PO}int \textbf{T}racking (\textbf{SPOT}) and real-time video processing. SPOT offers the following advantages:
\begin{itemize}
\item \emph{Strong Performance}: On CVO~\cite{wu2023accflow, le2024dense}, SPOT achieves state-of-the-art performance, outperforming the second one by 0.35 EPE (a 6.8\% error reduction) on Extended split. On TAP-Vid~\cite{doersch2022tap} and RoboTAP~\cite{vecerik2024robotap}, SPOT obtains up to 11.3\% relative improvement in AJ from the best online point tracking results.
\item \emph{Superior Efficiency}: SPOT tracks $512\times 512$ videos densely at 12.4 FPS on H100, consuming 4.15GB GPU memory. With 10$\times$ smaller parameter counts, it operates at least 2$\times$ faster than previous state-of-the-art models while still leading the accuracy on CVO~\cite{wu2023accflow, le2024dense}.
\item \emph{Causal Processing}: SPOT processes videos causally, on a frame-by-frame basis without relying on future frames, paving the way toward online dense motion perception.
\end{itemize}

SPOT consists of three main components: (1) a customized memory reading module based on the assumption of local affinity for feature enhancement; (2) a sensory memory for short-term motion dynamics modeling; (3) a visibility-guided splatting module for accurate long-range information propagation from the first frame. \cref{fig:teaser} shows a conceptual comparison with previous frameworks. 

The key idea in our approach is to break down the challenging task of long-range information propagation into two simpler steps that can be solved efficiently and accurately. First, we can use splatting~\cite{Niklaus_CVPR_2020} to transfer the features from the first frame to the coordinate grid of recent frames. With the accurate long-range optical flow already predicted, splatting ensures the features are mapped to correct positions for the same physical points across frames. Next, there is minimal appearance drifting between the recent frame and the current one, which enables using simple feature similarity (via an attention mechanism) to efficiently retrieve relevant features to enhance the current frame. As a result, regressing with the memory-enhanced feature can give us an accurate long-range optical flow.

The SPOT framework builds on ideas from existing research but introduces several key innovations. First, SPOT uses a memory bank to store distinctive features from the first frame and includes a specialized memory reading module to enhance the feature of current frame for accurate long-range optical flow prediction. In contrast, the recent memory-based model MemFlow~\cite{dong2024memflow} stores only short-term motion features and can only handle short-range optical flow between neighboring frames.

Second, SPOT includes an additional sensory memory to model short-term motion dynamics. This sensory memory, updated alternately with the flow, supports accurate long-range flow prediction. In comparison, recent sparse tracking methods~\cite{karaev2023cotracker, xiao2024spatialtracker} handle short-term motion by processing all frames within a sliding window, which is less efficient.

Third, SPOT’s regression operator uses a standard optical flow decoder, specifically the RAFT decoder~\cite{teed2020raft}. This approach unifies the architecture for both optical flow and point tracking, meaning any future improvements in optical flow architectures can directly enhance our model.

We conduct experiments on CVO~\cite{wu2023accflow, le2024dense}, TAP-Vid~\cite{doersch2022tap}, and RoboTAP~\cite{vecerik2024robotap}. The results demonstrate that SPOT achieves state-of-the-art performance for online tracking across these datasets. Additionally, we validate SPOT’s design choices through extensive ablation studies.

\section{Related Work}
\label{sec:related_work}

\begin{figure*}
\centering
\includegraphics[width=\linewidth]{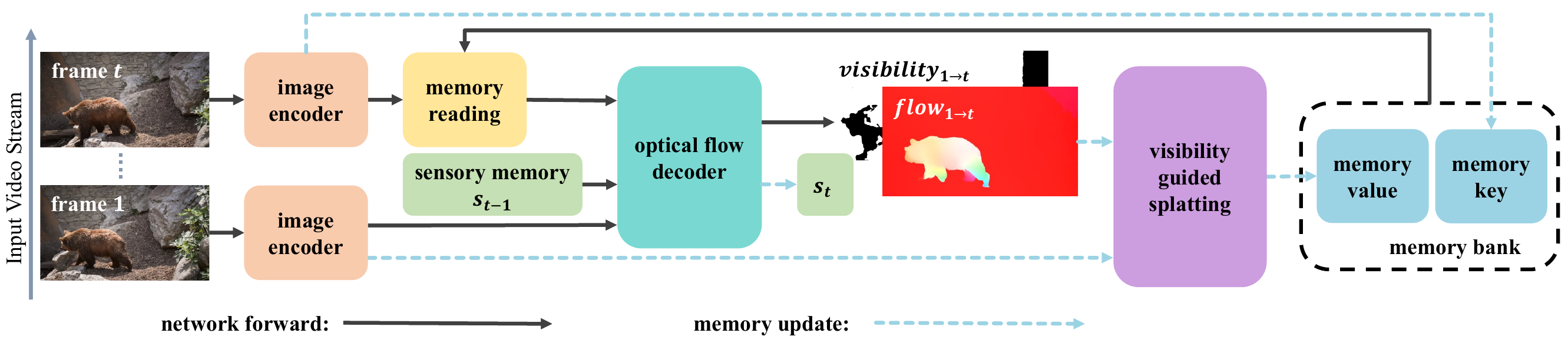}
\caption{Overview of our SPOT. With streaming video input, we first extract 4$\times$ downsampled feature for the current frame. Then the feature is enhanced by cross-attending to the memory to obtain a more discriminative feature. The optical flow and visibility mask are later obtained by a standard optical flow decoder with an additional input as sensory memory that models short-term motion dynamics. Finally, modulating by the visibility mask, we splat the feature of first frame to the memory value with the predicted flow for future usage. The memory key and sensory memory are updated by the current feature and the optical flow decoder, respectively.\label{fig:overview}}
\vspace{-0.15in}
\end{figure*}

\noindent\textbf{Optical Flow Estimation}. Optical flow estimation is traditionally formulated as an energy function minimization problem~\cite{horn1981determining, black1993framework, black1996robust, zach2007duality, revaud2015epicflow, brox2004high, bruhn2005lucas}, while recent works~\cite{dosovitskiy2015flownet, ilg2017flownet, sun2018pwc, hui2018liteflownet, ranjan2017optical, dong2023rethinking, xu2022gmflow, teed2020raft, jiang2021learning, zheng2022dip, sun2022skflow, huang2022flowformer, dong2024memflow} resort to deep learning network trained on large scale dataset for direct generation. However, these works typically focus on optical flow estimation between adjacent frames. Early attempts~\cite{lim2005optical, janai2017slow, crivelli2014robust, crivelli2012optical} have tried to chain short-range optical flow and deduce long-range motions. Recently, MFT~\cite{neoral2024mft} has improved the pipeline through reliable chaining identification, and AccFlow~\cite{wu2023accflow} introduces an occlusion robust backward accumulation strategy. Nonetheless, chaining-based methods have intrinsic limitations, \ie, forward accumulation~\cite{neoral2024mft} struggles with occlusion while backward accumulation~\cite{wu2023accflow} leads to linear increasing of processing time for a new frame. Our SPOT also builds upon the standard optical flow model~\cite{teed2020raft}. Yet distinct from chaining-based methods, SPOT employs an effective and efficient memory and shows it is indeed possible to achieve both highly accurate results and fast inference speed.

\noindent\textbf{Point Tracking}. Point Tracking aims to predict the position and visibility of the same physical point in every other video frame, given the query point in one frame.
Since the pioneering works, PIPs~\cite{harley2022particle} and TAP-Net~\cite{doersch2022tap}, point tracking has attracted a lot of attention and numerous works~\cite{zheng2023pointodyssey, doersch2023tapir, karaev2023cotracker, xiao2024spatialtracker, le2024dense, li2025taptr, vecerik2024robotap} have emerged. 
Among them, DOT~\cite{le2024dense} and Online TAPIR~\cite{vecerik2024robotap} tackle a similar task as ours. Specifically, DOT proposes an optical flow refinement stage for tracking densification. However, it depends on sparse tracks for initialization, which can easily miss small objects. Besides, combining several models also incurs a huge amount of model parameters. As for Online TAPIR, it proposes to replace the temporal convolutions with causal convolutions within TAPIR~\cite{doersch2023tapir} for online tracking, yet it is infeasible to track pixels densely in a reasonable amount of time. Conversely, SPOT estimates dense point tracking with both high accuracy and efficiency, achieved by a dedicated memory. Besides, leveraging real-world videos for self-supervised finetuning as in BootsTAP~\cite{doersch2024bootstap} is also a promising technique, although it is beyond the scope of our paper.

\noindent\textbf{Memory in Motion Estimation}. Memory mechanism has been widely used in various video tasks such as video object segmentation~\cite{oh2019video, cheng2021rethinking, hu2021learning, li2022recurrent, cheng2022xmem, cheng2024putting, ravi2024sam}, video semantic segmentation~\cite{wang2021temporal, paul2021local}, and video prediction~\cite{lee2021video}. However, VOS focuses on the identification of single or multiple object-level correspondences. It is non-trivial to directly extend them to dense and pixel-level correspondence tasks. Recently, MemFlow~\cite{dong2024memflow} pioneers using memory for optical flow estimation, but can only model short-range motion between adjacent frames. Importantly, our motivation originates from a deep analysis of online dense point tracking, which is naturally suitable for modeling with memory. Furthermore, we propose a novel splatting method for memory updating that is appropriate for long-range propagation.

\section{SPOT}
\label{sec:Methodology}

\subsection{Definition and Overview}
\noindent\textbf{Problem Setup}. The goal of point tracking is to predict the position $(x, y) \in \mathbb{R}^2$ and visibility $v \in [0, 1]$ of query point $(x^q, y^q) \in \mathbb{R}^2$ in other video frames, which is equivalent to predict the long-range optical flow $\mathbf{f}=(x - x^q, y - y ^q)$. In the setting of our online dense point tracking, we have access to all frames before the current timestamp $t$: $\{\mathbf{I}_1, \mathbf{I}_2, \cdots, \mathbf{I}_t\} \in \mathbb{R}^{t\times H\times W\times 3}$. We aim to predict the optical flow $\mathbf{f}_{1\rightarrow t} \in \mathbb{R}^{H\times W\times 2}$ and visibility mask $\mathbf{v}_{1\rightarrow t} \in [0, 1]^{H\times W}$ in a streaming fashion.

\noindent\textbf{Overview}. \cref{fig:overview} provides an overview of SPOT. First, SPOT encodes the current frame into 4$\times$ downsampled feature and conditions it on the memory bank to produce a more discriminative embedding (\cref{subsec:feature_enhancement}). Then optical flow decoder accepts enhanced feature, reference feature, and an additional sensory memory of shot-range motion dynamics as inputs, and outputs long-range optical flow with visibility mask (\cref{subsec:sensory_mem}). The sensory memory is also refreshed with internal motion feature within the flow decoder. SPOT finally updates the memory bank through visibility-guided splatting with predicted flow and reference feature for information propagation (\cref{subsec:mem_update}). In the following, we will describe each module sequentially and provide more implementation details in \cref{subsec:details}.

\subsection{Feature Enhancement with Memory Reading}
\label{subsec:feature_enhancement}

Given current video frame $\mathbf{I}_t$, we employ a standard residual convolutional network to extract dense feature at a lower resolution: $\mathbf{F}_t \in \mathbb{R}^{H/4\times W/4\times D}$, where $D$ is the feature dimension. As our memory bank consists of keys and values, the original feature $\mathbf{F}_t$ can serve as the query and the attention mechanism can be used for information retrieval. 

Specifically, the original feature $\mathbf{F}_t$ is projected through a linear projection layer for generating the query feature $q$. Then a dot-product similarity calculation is performed against the memory key followed by a softmax for normalization. This normalized score is used to aggregate the memory value to obtain the readout features $\mathbf{M}_t$. The process can be formulated as:
\begin{equation}
\mathbf{M}_t = \mathrm{Softmax}(1/\sqrt{D_k} \times q\times k^T)\times v,
\end{equation}
where $k \in \mathbb{R}^{L\times D_k}$ and $v \in \mathbb{R}^{L\times D_v}$ are keys and values with $L$ elements stored in the memory bank.

Yet, we would like to point out that it is inappropriate to use the readout features $\mathbf{M}_t$ for flow regression directly. As our memory value is obtained through splatting the reference feature $\mathbf{F}_1$ with the previously predicted flow (which will be introduced in \cref{subsec:mem_update}), there are many blank hole artifacts~\cite{Aleotti2021Learning, kar20223d} due to disocclusion~\cite{ilg2020estimating}. These disocclusion artifacts will hamper the network training and degrade the performance of flow regression. Inspired by the typically used inpainting technique for hole filling~\cite{Aleotti2021Learning}, we further propose a fusion layer for artifacts remedy. Fortunately, we find that one single convolution layer performs well enough, and we ``inpaint" the feature through the original feature $\mathbf{F}_t$:
\begin{equation}\label{equ:fusion}
{\mathbf{E}}_t = \mathbf{F}_t + \mathrm{Conv}(\mathbf{F}_t\oplus\mathbf{M}_t),
\end{equation}
where $\oplus$ denotes concatenation and ${\mathbf{E}}_t$ is the final enhanced feature that is ready for the next stage of flow regression.

\subsection{Flow Decoding and Sensory Memory}
\label{subsec:sensory_mem}
\begin{figure}
\centering
\includegraphics[width=0.9\linewidth]{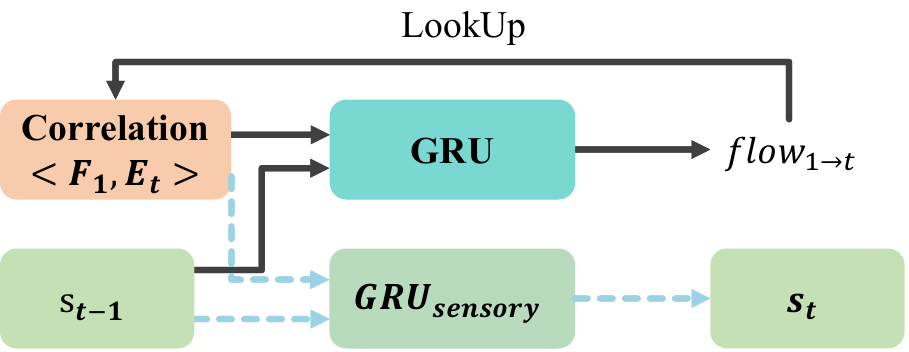}
\caption{Overview of \textbf{optical flow decoder} within \cref{fig:overview}. Motion feature is obtained through flow lookup. We update the flow with sensory memory, and update the sensory memory with the final motion feature. For the sake of simplicity, visibility prediction and context feature are omitted here.\label{fig:sensor_mem}}
\vspace{-0.15in}
\end{figure}

With the memory-enhanced feature ${\mathbf{E}}_t$ and the reference feature $\mathbf{F}_1$, we first compute the 4D correlation volume between all pairs following optical flow literature~\cite{teed2020raft}.
The motion feature $f_{m}^i$ is obtained through a lightweight motion encoder, taking current estimation $\mathbf{f}_{1\rightarrow t}^i$, $\mathbf{v}_{1\rightarrow t}^i$, and correlation volume as input. Finally, we produce an update direction for flow and visibility with a GRU unit as in \cite{le2024dense}:
\begin{equation}
\Delta f_{1\rightarrow t}^i, \Delta v_{1\rightarrow t}^i, h_t^i=\mathrm{GRU}(h_t^{i-1}, f_c, f_{m}^i, s_{t-1}),
\end{equation}
where $h_t^i$ is the hidden state of GRU, $f_c$ is the context feature of $\mathbf{I}_1$, and $s_{t-1}\in \mathbb{R}^{H/4\times W/4\times D_s}$ is the sensory memory of motion history. We perform a total of $N$ refinement steps of GRU and get the final optical flow, visibility mask, and corresponding motion feature $f_{m}^N$. Importantly, we update the sensory memory with final motion feature by an additional GRU for short-term motion dynamic modeling:
\begin{equation}
s_t={\mathrm{GRU}}_{sensory}(s_{t-1}, f_{m}^N).
\end{equation}
An overview of this process is illustrated in \cref{fig:sensor_mem}.

\subsection{Visibility-guided Splatting}
\label{subsec:mem_update}

The key to our formulation is to obtain discriminative features through memory reading. Recall that SPOT relies on feature similarity to aggregate memory value, while similarity score is only accurate within a handful of frames before the current frame due to appearance drifting. Therefore, we need to continually propagate the discriminative features of reference frame to latest frame. Fortunately, the predicted optical flow $\mathbf{f}_{1\rightarrow t}^N$ can be used for such long-range information propagation with both high accuracy and efficiency.

To achieve this, forward warping via splatting~\cite{Niklaus_CVPR_2020} becomes a natural choice. Specifically, for every position $(x_1, y_1)$, we calculate its position in the target frame by adding the optical flow. Then bilinear kernel $b(\cdot)$ (details in Supp.) is employed for weighted summation as follows:
\begin{align}\label{equ:summation_spla}
&F_t^\Sigma\left[(x_t, y_t)\right ] =\sum_{(x_1, y_1)} b(\Delta)\cdot \mathbf{F}_1\left[(x_1, y_1)\right ], \\
\text{where}\quad &\Delta = (x_1, y_1) + \mathbf{f}_{1\rightarrow t}^N\left[(x_1, y_1)\right ] - (x_t, y_t).
\end{align}
We denote the splatting results as $\sum\limits^{\rightarrow}(\mathbf{F}_1, \mathbf{f}_{1\rightarrow t})=F_t^\Sigma$. However, this splatting will lead to inconsistencies artificats~\cite{Niklaus_CVPR_2020} for occlusion regions. Therefore, we resort to our predicted visibility mask for occlusion handling and get the final splatting results as follows:
\begin{equation}\label{equ:splatting}
F_t^{1\rightarrow t}=\frac{\sum\limits^{\rightarrow}(\mathbf{v}_{1\rightarrow t}\cdot\mathbf{F}_1, \mathbf{f}_{1\rightarrow t})}{\sum\limits^{\rightarrow}(\mathbf{v}_{1\rightarrow t}, \mathbf{f}_{1\rightarrow t})},
\end{equation}
where we mask the invisible regions and normalize the results with visibility score.

Lastly, the memory bank maintains two FIFO queues of keys and values, caching a window of $L$ recent first-frame-derived features. We append the splatting results $F_t^{1\rightarrow t}$ into the queue of memory value. As for the memory key, we reuse the query feature $q$ of $\mathbf{F}_t$ for simplicity.

\subsection{Implementation Details}
\label{subsec:details}

Here, we describe more implementation details of SPOT.

\noindent\textbf{Warm-Start}. As SPOT is applied for streaming videos, we employ previous information to warm-start the estimation of next frame. Particularly, we initialize the hidden state $h_t^0$ of GRU with $h_{t-1}^N$ and initialize the flow $\mathbf{f}_{1\rightarrow t}^0$ by one step extrapolation: $\mathbf{f}_{1\rightarrow t}^0 = \mathbf{f}_{1\rightarrow t-1}^0 + 2\times(\mathbf{f}_{1\rightarrow t-1}^N - \mathbf{f}_{1\rightarrow t-1}^0)$.

\noindent\textbf{Training}. The loss function of SPOT consists of the $l_1$ loss~\cite{teed2020raft} for the predicted flows and binary cross-entropy loss for the visibility prediction. SPOT is first trained on synthetic optical flow dataset, \ie, Kubric-CVO~\cite{wu2023accflow}, for 500k steps at resolution $384\times 384$. Then it is finetuned on Kubric-MOVi-F~\cite{greff2022kubric} for 100k steps with 24-frame videos of resolution $384\times 384$. During training, we set the iteration of flow decoding $N$ to 12. If not otherwise mentioned, we set $N$ to 16 during inference. To reduce the drifting effect, we set the memory bank length $L$ to 3.

\section{Experiments}
\label{sec:Experiments}

\begin{table*} \small
\centering
\caption{Long-range optical flow estimation on CVO. We evaluate dense optical flow predictions from the first to the last frame of videos. DOT~\cite{le2024dense} here uses CoTracker2~\cite{karaev2023cotracker} for the first stage of sparse point tracking. {*} uses the flow of neighboring frames as an initialization for more distant frames. DOT$^{\dagger}$ represents evaluated the first stage of DOT~\cite{le2024dense} in an online fashion with sliding windows of 2.\label{tab:quatitative_cvo}}
\begin{tabular}{ccccccc}
\toprule
\multirow{2}{*}{Method} & \multicolumn{2}{c}{CVO (Clean)} & \multicolumn{2}{c}{CVO (Final)} & \multicolumn{2}{c}{CVO (Extended)}\tabularnewline
\cmidrule(lr){2-3} \cmidrule(lr){4-5} \cmidrule(lr){6-7}
 & EPE $\downarrow$ (all / vis / occ) & OA $\uparrow$ & EPE $\downarrow$ (all / vis / occ) & OA $\uparrow$ & EPE $\downarrow$ (all / vis / occ) & OA $\uparrow$\tabularnewline
\midrule
\textbf{Offline} &  &  &  &  &  & \tabularnewline
PIPs++~\cite{zheng2023pointodyssey} & 9.05 / 6.62 / 21,5 & - & 9.49 / 7.06 / 22.0 & - & 18.4 / 10.0 / 32.1 & -\tabularnewline
TAPIR~\cite{doersch2023tapir} & 4.15 / 1.48 / 16.5 & 96.9 & 4.59 / 1.88 / 17.2 & 96.6 & 22.6 / 4.70 / 49.7 & 88.3\tabularnewline
CoTracker2~\cite{karaev2023cotracker} & 1.50 / 0.90 / 4.41 & 97.1 & 1.47 / 0.94 / 4.20 & 97.0 & 5.45 / 4.03 / 8.30 & 88.2\tabularnewline
SpatialTracker~\cite{xiao2024spatialtracker} & 1.84 / 1.32 / 4.72 & - & 1.88 / 1.37 / 4.68 & - & 5.53 / 4.18 / 8.68 & -\tabularnewline
DOT~\cite{le2024dense} & \textbf{1.34 / 0.76 / 4.11} & \textbf{97.6} & \textbf{1.37 / 0.84 / 4.06} & \textbf{97.5} & \textbf{5.12 / 3.65 / 7.61} & \textbf{88.9}\tabularnewline
\midrule
\textbf{Online} &  &  &  &  &  & \tabularnewline
RAFT~\cite{teed2020raft} & 2.82 / 1.70 / 8.01 & 90.2 & 2.88 / 1.79 / 7.89 & 89.7 & 28.6 / 21.6 / 41.0 & 62.3\tabularnewline
GMA~\cite{jiang2021learning} & 2.90 / 1.91 / 7.63 & 91.0 & 2.92 / 1.89 / 7.48 & 90.6 & 30.0 / 22.8 / 42.6 & 62.0\tabularnewline
RAFT{*}~\cite{teed2020raft} & 2.19 / 1.20 / 6.77 & 90.1 & 2.35 / 1.35 / 6.76 & 89.6 & 21.1 / 23.0 / 25.6 & 63.6\tabularnewline
GMA{*}~\cite{jiang2021learning} & 2.16 / 1.19 / 6.59 & 91.1 & 2.25 / 1.28 / 6.53 & 90.8 & 21.0 / 22.4 / 25.2 & 64.0\tabularnewline
MFT~\cite{neoral2024mft} & 2.91 / 1.39 / 9.93 & 95.8 & 3.16 / 1.56 / 10.3 & 95.7 & 21.4 / 9.20 / 41.8 & 85.5\tabularnewline
AccFlow~\cite{wu2023accflow} & 1.69 / 1.08 / 4.70 & 84.8 & 1.73 / 1.15 / 4.63 & 84.4 & 36.7 / 28.1 / 52.9 & 58.8\tabularnewline
DOT$^{\dagger}$~\cite{le2024dense} & 1.92 / 1.07 / 6.10 & \textbf{97.1} & 1.98 / 1.17 / 6.02 & \textbf{97.1} & 12.1 / 7.06 / 20.7 & 87.0\tabularnewline
\textbf{SPOT (Ours)} & \textbf{1.11 / 0.60 / 3.61} & 96.8 & \textbf{1.23 / 0.72 / 3.67} & 96.5 & \textbf{4.77 / 3.58 / 6.96} & \textbf{88.6}\tabularnewline
\bottomrule
\end{tabular}
\vspace{-0.15in}
\end{table*}

We conduct evaluations on three benchmarks: long-range optical flow benchmark CVO~\cite{wu2023accflow, le2024dense}, sparse point tracking benchmark TAP-Vid~\cite{doersch2022tap}, and RoboTAP~\cite{vecerik2024robotap}. SPOT is mainly compared with recent online methods, \eg MFT~\cite{neoral2024mft} and Online TAPIR~\cite{vecerik2024robotap}. We also consider several offline methods for comparison, namely CoTracker2~\cite{karaev2023cotracker}, TAPTR~\cite{li2025taptr}, and DOT~\cite{le2024dense}. We further provide a more comparable version of DOT by reducing their sliding window size to 2 (DOT$^\dagger$) for an online dense point tracking evaluation, which is the most similar protocol with reasonable inference speed we pursue here. We present the evaluation protocols and comparison results as follows.

\subsection{CVO}

CVO test set~\cite{wu2023accflow} originally includes Clean and Final pass, each containing 536 videos of 7 frames at 60 FPS. Lately, DOT~\cite{le2024dense} introduces the Extended set with another 500 videos of 48 frames at 24 FPS. All videos are provided with groundtruth optical flow and visibility mask between first and last frame. We utilize end-point-error (EPE) within all, visible, and occluded regions for flow evaluation, and occlusion accuracy (OA) for visibility prediction evaluation.

\begin{figure*}
\centering
\includegraphics[width=0.8\linewidth]{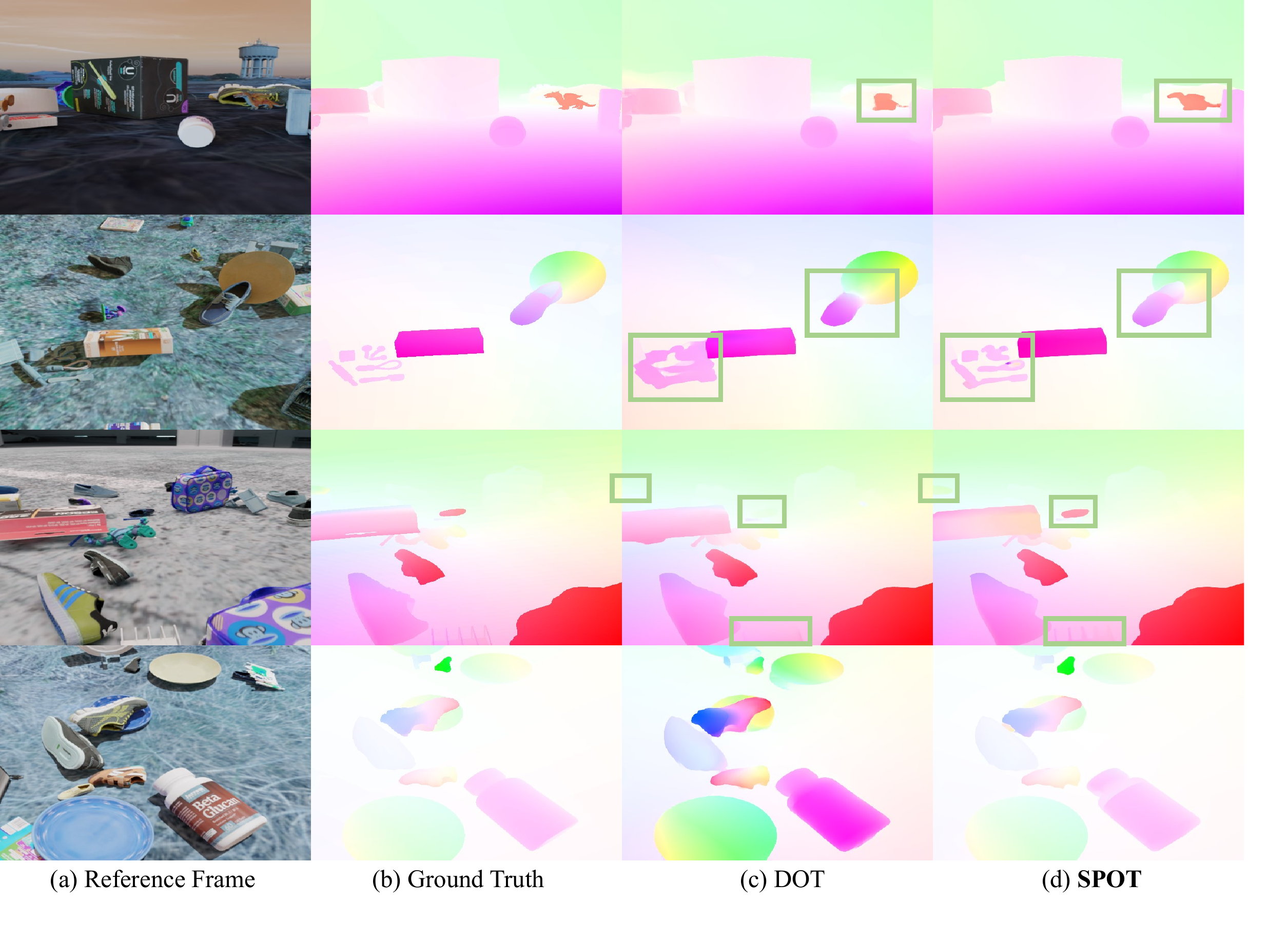}
\caption{Qualitative comparison on the CVO (Extended). Notable areas are marked by a bounding box. Please zoom in for details.\label{fig:qualitative_cvo}}
\end{figure*}

As shown in \cref{tab:quatitative_cvo}, SPOT achieves state-of-the-art performance on all three CVO test sets in terms of EPE. Particularly, SPOT achieves a 60.5\% reduction from 12.08 to 4.77 compared to the online DOT$^\dagger$ on the Extended set. It even outperforms all offline methods across three sets for flow prediction. This is due to CVO only evaluating the flow between the first and last frame, there is no future frame that can be utilized by offline methods to improve the performance. This demonstrates that SPOT enables the exploitation of past information more effectively. Besides, SPOT outperforms MFT in terms of occlusion accuracy and performs comparably with DOT$^\dagger$. \cref{fig:qualitative_cvo} provides a qualitative comparison with DOT, SPOT successfully predicts the flow for small objects and suffers less drifting than DOT.

\subsection{TAP-Vid and RoboTAP}

\begin{table*} \small
\setlength{\tabcolsep}{2.5pt}
\centering
\caption{Point tracking results on TAP-Vid and RoboTAP. DOT$^{\dagger}$ represents evaluated in an online fashion with sliding windows of 2.\label{tab:quatitative_davis}}
\begin{tabular}{cccccccccccccccc}
\toprule
\multirow{2}{*}{Method} & \multicolumn{3}{c}{DAVIS (First)} & \multicolumn{3}{c}{DAVIS (Strided)} & \multicolumn{3}{c}{RGB-S. (First)} & \multicolumn{3}{c}{RoboTAP (First)} & \multicolumn{3}{c}{Kinetics (First)}\tabularnewline
\cmidrule(lr){2-4} \cmidrule(lr){5-7} \cmidrule(lr){8-10}  \cmidrule(lr){11-13} \cmidrule(lr){14-16}
 & AJ $\uparrow$ & $<\delta_{avg}^{x}\uparrow$ & OA $\uparrow$ & AJ $\uparrow$ & $<\delta_{avg}^{x}\uparrow$ & OA $\uparrow$ & AJ $\uparrow$ & $<\delta_{avg}^{x}\uparrow$ & OA $\uparrow$ & AJ $\uparrow$ & $<\delta_{avg}^{x}\uparrow$ & OA $\uparrow$ & AJ $\uparrow$ & $<\delta_{avg}^{x}\uparrow$ & OA $\uparrow$\tabularnewline
\midrule
\textbf{Offline} &  &  &  &  &  &  &  &  &  &  &  &  &  &  & \tabularnewline
PIPs~\cite{harley2022particle} & 42.2 & 64.8 & 77.7 & 52.4 & 70.0 & 83.6 & - & - & - & 29.5 & - & - & 31.7 & 53.7 & 72.9\tabularnewline
TAPIR~\cite{doersch2023tapir} & 56.2 & 70.0 & 86.5 & 61.3 & 73.6 & 88.8 & 55.5 & 69.7 & 88.0 & 59.6 & 73.4 & 87.0 & 49.6 & 64.2 & 85.0\tabularnewline
CoTracker2~\cite{karaev2023cotracker} & 60.8 & 74.8 & 88.4 & 65.9 & 79.4 & 89.9 & 60.5 & 73.3 & 83.5 & 58.6 & 70.6 & 87.0 & 48.4 & 62.2 & 83.2\tabularnewline
SpatialTracker~\cite{xiao2024spatialtracker} & 61.1 & \textbf{76.3} & 89.5 & - & - & - & 63.5 & 77.6 & 88.2 & - & - & - & \textbf{50.1} & \textbf{65.9} & \textbf{86.9}\tabularnewline
TAPTR~\cite{li2025taptr} & \textbf{63.0} & 76.1 & \textbf{91.1} & 66.3 & 79.2 & 91.0 & 60.8 & 76.2 & 87.0 & 60.1 & \textbf{75.3} & 86.9 & 49.0 & 64.4 & 85.2\tabularnewline
DOT~\cite{le2024dense} & 62.2 & 76.1 & 89.9 & \textbf{67.8} & \textbf{80.6} & \textbf{91.1} & \textbf{77.5} & \textbf{87.4} & \textbf{92.6} & \textbf{61.5} & 73.7 & \textbf{87.9} & 48.4 & 63.8 & 85.2\tabularnewline
\midrule
\textbf{Online} &  &  &  &  &  &  &  &  &  &  &  &  &  &  & \tabularnewline
TAP-Net~\cite{doersch2022tap} & 33.0 & 48.6 & 78.8 & 38.4 & 53.1 & 82.3 & 53.5 & 68.1 & 86.3 & 45.1 & - & - & 38.5 & 54.4 & 80.6\tabularnewline
MFT~\cite{neoral2024mft} & 47.3 & 66.8 & 77.8 & 56.1 & 70.8 & 86.9 & - & - & - & - & - & - & 39.6 & 60.4 & 72.7\tabularnewline
DOT$^{\dagger}$~\cite{le2024dense} & 53.3 & 67.8 & 85.4 & 60.1 & 73.8 & 87.3 & 61.3 & 75.3 & 86.5 & 51.9 & 62.9 & 79.9 & 45.3 & 58.0 & 81.4\tabularnewline
Online TAPIR~\cite{vecerik2024robotap} & 56.2 & 69.3 & 84.6 & 58.3 & 70.6 & 84.6 & 65.9 & 78.6 & 89.5 & 57.9 & 69.9 & 86.1 & 49.6 & 61.1 & 83.1\tabularnewline
\textbf{SPOT (Ours)} & \textbf{61.5} & \textbf{75.0} & \textbf{88.9} & \textbf{66.9} & \textbf{79.4} & \textbf{90.2} & \textbf{73.3} & \textbf{86.4} & \textbf{90.3} & \textbf{60.7} & \textbf{73.6} & \textbf{87.5} & \textbf{50.2} & \textbf{62.7} & \textbf{85.5}\tabularnewline
\bottomrule
\end{tabular}
\vspace{-0.15in}
\end{table*}

TAP-Vid~\cite{doersch2022tap} provides 3 datasets with different scenarios: 30 real-world videos from DAVIS~\cite{perazzi2016benchmark}, over 1,000 human actions videos from Kinetics~\cite{carreira2017quo}, and 50 synthetic videos for robotic manipulation from RGB-Stacking~\cite{lee2021beyond}. In addition, RoboTAP~\cite{vecerik2024robotap} further provides a dataset with 265 real-world videos of robotic manipulation tasks. We evaluate SPOT on DAVIS in both ``first” and ``strided" query modes~\cite{doersch2022tap} while evaluating other datasets in ``first” mode solely. Following the standard evaluation protocol of TAP-Vid, we first downsample the videos to $256\times 256$ and report occlusion accuracy (OA), position accuracy ($<\delta_{avg}^{x}$) for visible points, and average jaccard (AJ) which evaluates both occlusion and position accuracy.

The quantitative comparisons are reported in \cref{tab:quatitative_davis}, which shows SPOT consistently outperforms other online methods on all datasets. Specifically, when compared to previous state-of-the-art online tracking models, the relative improvements of SPOT up to 11.3\% in average jaccard (AJ), 9.9\% in position accuracy ($<\delta_{avg}^{x}$), and 4.1\% in occlusion accuracy (OA). SPOT even outperforms some recent offline tracking methods slightly, \eg, beats CoTracker2~\cite{karaev2023cotracker} on all datasets and performs better than SpatialTracker~\cite{xiao2024spatialtracker} in terms of AJ. Furthermore, SPOT achieves the best AJ in Kinetics.

\begin{figure*}
\centering
\includegraphics[width=0.9\linewidth]{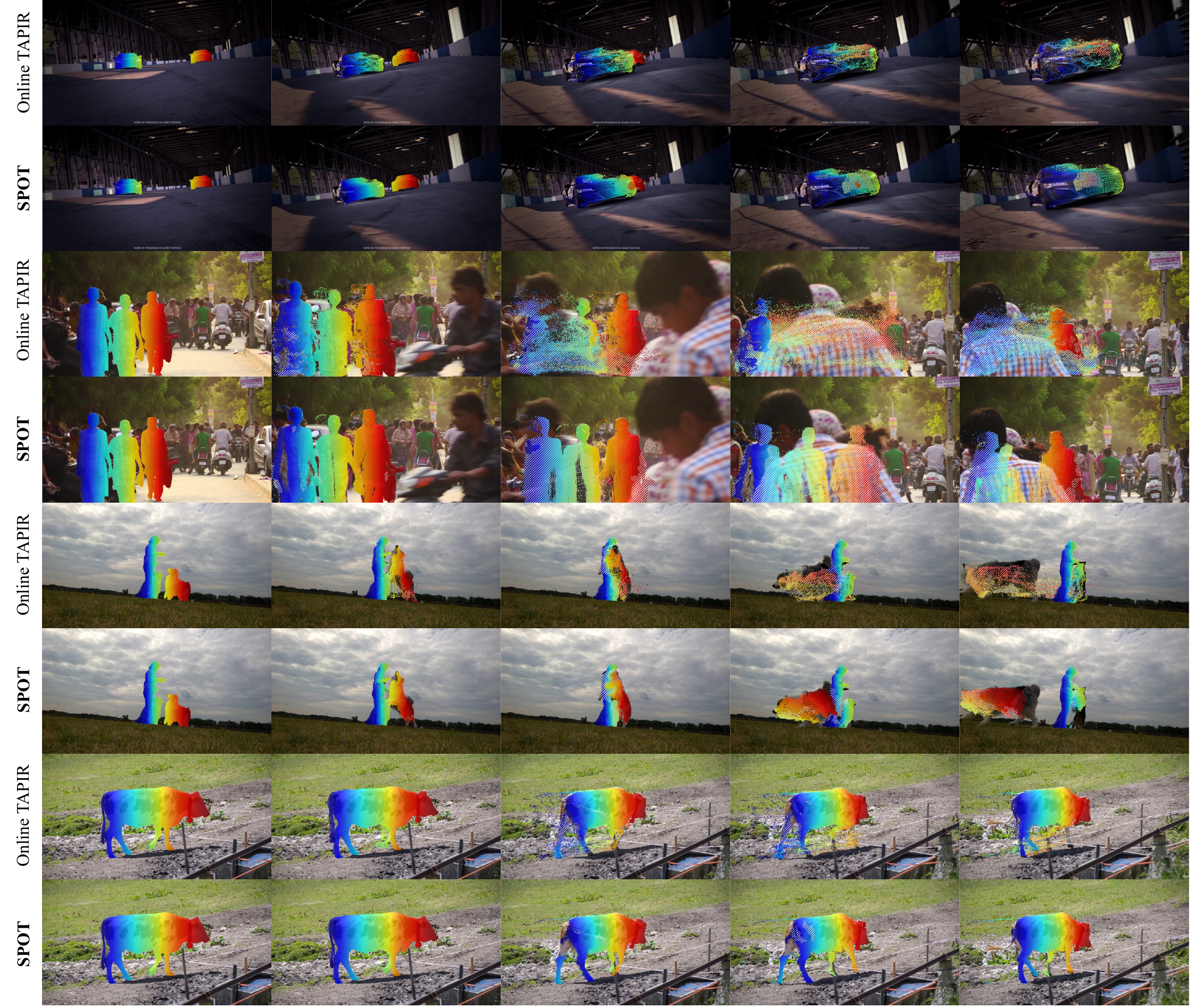}
\caption{Qualitative comparison on DAVIS. For each sequence, we show tracking results of Online TAPIR~\cite{vecerik2024robotap} and SPOT. Only foreground points of the first frame are visualized, each point is displayed with a different color and overlayed with white stripes if occluded.\label{fig:qualitative_davis}}
\vspace{-0.15in}
\end{figure*}

We further provide qualitative comparisons with Online TAPIR~\cite{vecerik2024robotap}, which is currently the best method for online tracking, on challenging videos from DAVIS in \cref{fig:qualitative_davis}. SPOT outperforms Online TAPIR in handling complex occlusion, estimating better and consistent motion state of the car and people behind distractors (the first and second examples of \cref{fig:qualitative_davis}), achieves better tracking during multi-subjects interaction (the third example), and effectively tracks small part of the animal (the fourth example). Animated video results can be found in our project page.

\subsection{Timing and Parameter Counts}

\begin{figure}
\centering
\includegraphics[width=\linewidth]{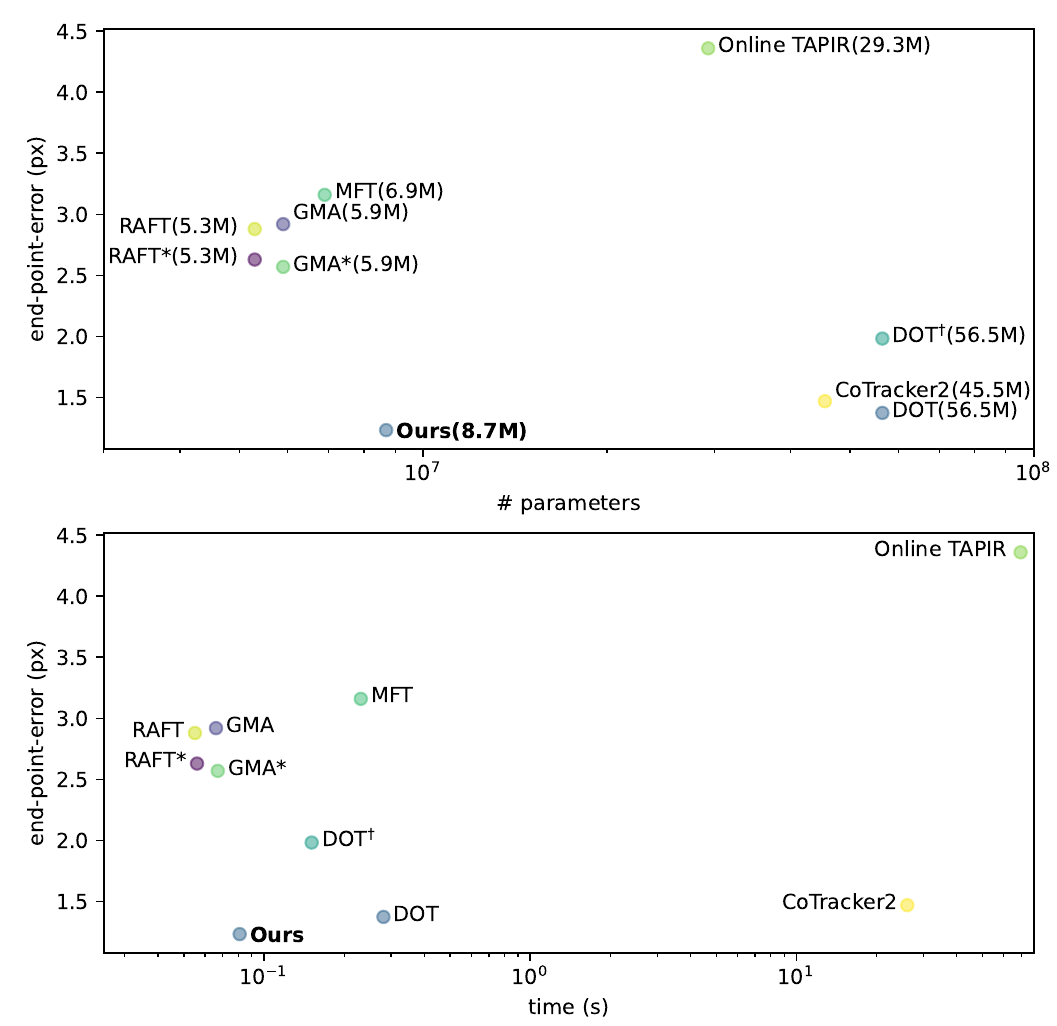}
\vspace{-0.3in}
\caption{Plots comparing parameter counts (M) and inference time (s) vs. accuracy. Accuracy is measured by the EPE on the CVO (Final) test set. \emph{Top}: Parameter count vs. accuracy compared to other methods. Our method is more parameter efficient while achieving lower EPE than previous state-of-the-art methods. \emph{Bottom}: Inference time per frame of 512$\times$512 resolution vs. accuracy on one NVIDIA H100 GPU.\label{fig:epe_time_para}}
\vspace{-0.1in}
\end{figure}

We present inference time and parameter counts in \cref{fig:epe_time_para}. Accuracy is determined by performance on the CVO (Final) set. The parameter number of SPOT is only 8.7M, which is an order of magnitude smaller than the number of parameters of other point tracking methods, \eg, 29.3M for Online TAPIR and 56.5M for DOT. For video input of $512\times 512$, SPOT can track all points within the first frame and runs at 12.4 FPS on one H100 GPU, which is much faster than sparse tracking methods like Online TAPIR and CoTracker2. And SPOT also runs faster than DOT and DOT$^\dagger$ in our hardware without sacrificing performance. Besides, our parameter number and inference speed are similar to recent optical flow models while performing much better than them in terms of dense point tracking.

\subsection{Ablation study}

In this section, we perform a series of ablation studies on SPOT to show the relative importance of each module. All experiments are conducted on CVO (Extended). For time-saving, SPOT is trained on 10-frame videos in \cref{tab:ablation}.

\noindent\textbf{Module Ablation}. We first experiment with the importance of each network module as in \cref{tab:ablation}. Specifically, we first remove the feature fusion module (\cref{equ:fusion}) during the stage of feature enhancement. Without feature fusion, the blank hole artifacts will soon take over the network and easily lead to NaN during training. Furthermore, removing the memory bank and memory reading together results in a classical optical flow estimation model. Though trained on point tracking dataset Kubric-MOVi-F~\cite{greff2022kubric}, the model without memory bank performs much worse than the full model, which confirms the contribution of memory bank. Also, the results benefit from the sensory memory, which makes the model aware of short-term motion dynamics and improves the EPE from 8.64 to 6.42. Besides, an additional query projector for memory reading can improve the results slightly.

\noindent\textbf{Splatting Type}. We then present the ablation of different splatting strategies as in the second part of \cref{tab:ablation}. The splatting strategy used in \cref{equ:splatting} is dubbed as linear splatting in \cite{Niklaus_CVPR_2020}. Average splatting uses all one tensor for mask and normalization, while summation splatting corresponds to \cref{equ:summation_spla}.
The softmax splatting proposed in \cite{Niklaus_CVPR_2020} uses additional exponential functions for mask calculation. As shown in \cref{tab:ablation}, the linear splatting achieves the best performance, outperforming the second one, softmax, by 0.62 EPE.

\noindent\textbf{Warm-Start}. We finally validate the effectiveness of warm-start strategy in the last part of \cref{tab:ablation}. Initializing the hidden state $h_t^0$ of GRU with $h_{t-1}^N$ indeed results in significant performance gains, improving the EPE by 23.7\%. This may be because the hidden state also carries the motion information from the previous frame. Besides, warm-start the flow also improves the performance slightly.

\begin{table} \small
\setlength{\tabcolsep}{2.5pt}
\centering
\caption{Ablation studies on CVO (Extended). Settings used in our final model are underlined. We use 10-frame videos for training.\label{tab:ablation}}
\vspace{-0.1in}
\begin{tabular}{cccc}
\toprule
Experiment & Method & EPE $\downarrow$ (all / vis / occ) & OA $\uparrow$\tabularnewline
\hline
\multirow{5}{*}{Module Ablation} & \uline{Full} & \textbf{6.42 / 3.86 / 9.98} & \textbf{88.5}\tabularnewline
 & - Feature fusion & NaN & NaN\tabularnewline
 & - Memory bank & 38.98 / 28.34 / 57.15 & 78.8\tabularnewline
 & - Sensory memory & 8.64 / 4.55 / 14.60 & 88.2\tabularnewline
 & - Query projector & 6.48 / 3.88 / 9.82 & 88.3\tabularnewline
\midrule
\multirow{4}{*}{Splatting Type} & \uline{Linear} & \textbf{6.42 / 3.86 / 9.98} & \textbf{88.5}\tabularnewline
 & Average & 7.34 / 3.94 / 11.41 & 88.5\tabularnewline
 & Softmax & 7.04 / 4.35 / 10.60 & 88.5\tabularnewline
 & Summation & 7.17 / 3.96 / 11.17 & 88.5\tabularnewline
\midrule
\multirow{3}{*}{Warm-Start} & \uline{Full} & \textbf{6.42 / 3.86 / 9.98} & \textbf{88.5}\tabularnewline
 & - Hidden state & 7.94 / 3.81 / 13.17 & 88.3\tabularnewline
 & - Flow & 6.49 / 3.92 / 10.05 & 88.4\tabularnewline
\bottomrule
\end{tabular}
\vspace{-0.15in}
\end{table}

\noindent\textbf{Training Video Length}. We finally probe the importance of training video length to the final performance, as demonstrated in \cref{tab:ablation_length}. We experiment with video length (7, 10, and 24) during training and report the performance on CVO (Extended). Generally, we find that increasing the video length of training can improve the model performance. However, it also brings a huge memory consumption burden to our GPU hardware during training. Therefore, we finally choose to use 24-frame videos for training, which just fit into a NVIDIA 80GB H100 GPU.

\begin{table}\small
\setlength{\tabcolsep}{2.5pt}
\centering
\caption{Ablations of video/memory length on CVO (Extended).\label{tab:ablation_length}}
\vspace{-0.1in}
\begin{tabular}{cccc}
\toprule
Experiment & Length & EPE $\downarrow$ (all / vis / occ) & OA $\uparrow$\tabularnewline
\hline
\multirow{3}{*}{Video Length} & 7 & 7.14 / 3.94 / 11.56 & 88.5\tabularnewline
 & 10 & 6.42 / 3.86 / 9.98 & 88.5\tabularnewline
 & \uline{24} & \textbf{4.77 / 3.58 / 6.96} & \textbf{88.6}\tabularnewline
\midrule
\multirow{3}{*}{Memory Length} & 1 & 5.00 / 3.75 / 7.18 & 88.2\tabularnewline
 & \uline{3} & \textbf{4.77 / 3.58 / 6.96} & \textbf{88.6}\tabularnewline
 & 6 & 5.06 / 3.73 / 7.37 & 88.0\tabularnewline
\bottomrule
\end{tabular}
\vspace{-0.15in}
\end{table}

\noindent\textbf{Discussion}. We have demonstrated empirically that online dense point tracking can be resolved effectively by the memory mechanism. The intuition is that discriminative information can be propagated from the first frame to the latest frame within memory via already predicted accurate optical flow effectively, and this information can be readout by the attention mechanism based on feature similarity. As there is typically one or two timesteps between the current frame and the frame within memory, the feature similarity is accurate enough for information retrieval. This two-step information propagation mechanism is the cornerstone of the success of our approach. 

\noindent\textbf{Limitation}. However, the success of two-step propagation depends on the success of each step. Therefore, for accurate attention-based readout without suffering from the drifting problem, the length of the memory bank is typically small (3 in SPOT as verified in \cref{tab:ablation_length}). If there is a video with a extreme long time of occlusion, then SPOT may degrades to pairwise method, which is the main limitation of SPOT. To deal with such scenarios, one possible future work is to introduce persistent object-level memory, for the objects within the first frame, that updates adaptively and survives through the long occlusion. Besides, our SPOT is also not robust to extreme fast motion and texture-less region. And we provide more failure cases in Supplementary.

\section{Conclusion}
\label{sec:Conclusion}

We introduced SPOT, a new framework designed for efficient and accurate online dense point tracking. At the core of SPOT is a memory-based approach that enables effective information propagation across frames, supported by our feature enhancement and visibility-guided splatting modules. Additionally, SPOT’s sensory memory efficiently models short-term motion dynamics. SPOT achieves state-of-the-art results on CVO, outperforming other online methods and matching the accuracy of recent offline models on TAP-Vid and RoboTAP. With its low parameter count and fast processing speed, SPOT brings us closer to real-time, frame-by-frame motion tracking, opening up new possibilities for practical applications.

{\section*{Acknowledgements} 
This work was sponsored by Doubao Fund, ByteDance. Yanwei Fu  is also with Shanghai Innovation Institute, Institute of Trustworthy Embodied AI, Fudan University,  and ISTBI–ZJNU Algorithm Centre for Brain-inspired Intelligence, Zhejiang Normal University.}

{
    \small
    \bibliographystyle{ieeenat_fullname}
    \bibliography{main}
}

\clearpage
\setcounter{page}{1}
\maketitlesupplementary

\section{Implementation Details}

\noindent\textbf{Network Details}. We adopt the classical optical flow estimation network RAFT~\cite{teed2020raft} as our backbone. Besides, following the modification of DOT~\cite{le2024dense} to the architecture of RAFT, we also use a stride 1 in the first convolutional layer of the image encoder and predict the visibility mask by an additional mask decoder. The channel number of different features is set as follows: $D$ of encoded feature from the image encoder is 256, $D_k$ of memory key is 128, $D_v$ of memory value is 256, and $D_s$ of sensory memory is 128. In addition, the bilinear kernel $b(\cdot)$ used in our visibility-guided splatting has the following formulation:
\begin{equation}
b(\Delta)=\max(0, 1-|\Delta_x|)\cdot\max(0, 1-|\Delta_y|),
\end{equation}
where $\Delta = (x_1, y_1) + \mathbf{f}_{1\rightarrow t}^N\left[(x_1, y_1)\right ] - (x_t, y_t)$. 

\noindent\textbf{Loss Function}. The loss function of SPOT consists of the $l_1$ loss~\cite{teed2020raft} for the predicted flows and binary cross-entropy loss for the visibility prediction. Specifically, we use exponentially increasing weights for predictions from different GRU iterations. Given ground-truth optical flow $\mathbf{f}_{1\rightarrow t}^{gt}$ and visibility mask $\mathbf{v}_{1\rightarrow t}^{gt}$, our loss function is defined as:
\begin{equation}\notag
\sum_{i=1}^N0.8^{N-i}\left[\lambda||\mathbf{f}_{1\rightarrow t}^{gt}-\mathbf{f}_{1\rightarrow t}^i||_1 + \mathrm{BCE}(\mathbf{v}_{1\rightarrow t}^{gt}, \mathbf{v}_{1\rightarrow t}^i)\right],
\end{equation}
where $\lambda$ is set to 1000 empirically.

\noindent\textbf{Training Details}. We employ Flash{A}ttention-2~\cite{dao2023flashattention2} for fast attention computation within the memory reading module. During training, we use Adam optimizer with one-cycle~\cite{smith2019super} learning rate on eight NVIDIA H100 GPUs. The learning rate is 1e-4. The batch size is 24 for the first training stage (\ie, 500k steps on optical flow dataset Kubric-CVO~\cite{wu2023accflow}) and 8 for the second one (\ie, 100k steps on point tracking dataset Kubric-MOVi-F~\cite{greff2022kubric}).

\noindent\textbf{Evaluation Details}. We set the iteration number $N$ of flow decoding to 16 by default, as 16 iterations already achieve the peak accuracy on real-world videos from DAVIS~\cite{perazzi2016benchmark}. In contrast, we set it to 32 on Kinetics~\cite{carreira2017quo}, RGB-Stacking~\cite{lee2021beyond}, and RoboTAP~\cite{vecerik2024robotap}. Because we find that more iterations of GRU can improve the accuracy of these three datasets. These may be due to the different motion characteristics (\eg, faster motion of human actions and weak texture of robotic scenes) of these three datasets.

\section{More Quantitative Analysis}

\noindent\textbf{Online Setup under Original Window Size}. We additionally give a setup without modifying the window size of existing offline models: when processing frame $t$, we provide the model with all prior frames. This process is repeated for each frame sequentially, ensuring predictions are based solely on past frames. This gives an upper bound of performance for reference though with extremely slow speed. We use $\ddagger$ to denote this setup. Due to the extremely slow inference speed of this setup, we only provide the result on DAVIS (First) in \cref{tab:davis_online_evaluation}: SPOT still beats other online versions by a large margin, with a much faster inference speed. It still supports our contributions of strong performance and superior efficiency.

\begin{table}
\centering
\caption{Point tracking results on DAVIS (First). $\ddagger$ represents evaluated in an online fashion under the original window size with extremely slow speed.\label{tab:davis_online_evaluation}}
\begin{tabular}{cccc}
\toprule
\textbf{Online} & AJ $\uparrow$ & $<\delta_{avg}^{x}\uparrow$ & OA $\uparrow$\tabularnewline
\midrule
TAPIR$^{\ddagger}$ & 56.7 & 70.2 & 85.7\tabularnewline
CoTracker2$^{\ddagger}$ & 55.9 & 68.7 & 83.7\tabularnewline
SpatialTracker$^{\ddagger}$ & 57.3 & 70.6 & 85.0\tabularnewline
DOT$^{\ddagger}$ & 57.3 & 69.7 & 85.2\tabularnewline
Online TAPIR & 56.2 & 69.3 & 84.6\tabularnewline
\textbf{Ours} & \textbf{61.5} & \textbf{75.0} & \textbf{88.9}\tabularnewline
\bottomrule
\end{tabular}
\end{table}

\noindent\textbf{Further Discussion on Forward Splatting and Backward Warping}. The main idea of SPOT is breaking down long-term tracking into two simpler steps, i.e., long-range propagation with flow and similarity-based short-range retrieval. SPOT instantiates propagation with splatting and retrieval with attention. We can also first use attention to algin $\mathbf{F}_t$ to memory frames, then backward warp them to frame 1, i.e., a reversed pipeline of SPOT. Though there are many-to-one mapping and empty regions in splatting, we tackle them with visibility mask and inpainting. However, there are also problems with backward warping, i.e., warping error foreground for occluded region, warping out-of-frame empty regions. These problems are caused by underlying physical motion. Therefore, there is no preference for which long-range propagation method to choose. The important point is the framework of two-step breakdown. Here, we further provide an additional ablation experiment: we warp the features of memory frames to the first frame directly, without the stage of attention for short-range retrieval. As shown in \cref{tab:two_step_abla}, though the ablated model achieve similar results as our SPOT on short videos, it fails to generalize to longer videos (CVO Extended) due to error accumulation.

\begin{table*}
\centering
\caption{Alation results on CVO dataset. We ablate the attention block of SPOT here. \label{tab:two_step_abla}}
\begin{tabular}{ccc}
\toprule
\multirow{2}{*}{Method} & \multicolumn{1}{c}{CVO (Final)} & \multicolumn{1}{c}{CVO (Extended)}\tabularnewline
\cmidrule{2-2}\cmidrule{3-3}
 & EPE $\downarrow$ (all / vis / occ) & EPE $\downarrow$ (all / vis / occ)\tabularnewline
\midrule
\textbf{SPOT} & \textbf{1.17 / 0.67 / 3.49} & \textbf{6.42 / 3.86 / 9.98}\tabularnewline
Attention Ablation & 1.18 / 0.69 / 3.50 & 395.35 / 379.44 / 428.48
\tabularnewline
\bottomrule
\end{tabular}
\end{table*}

\begin{figure}
\centering
\includegraphics[width=0.8\linewidth]{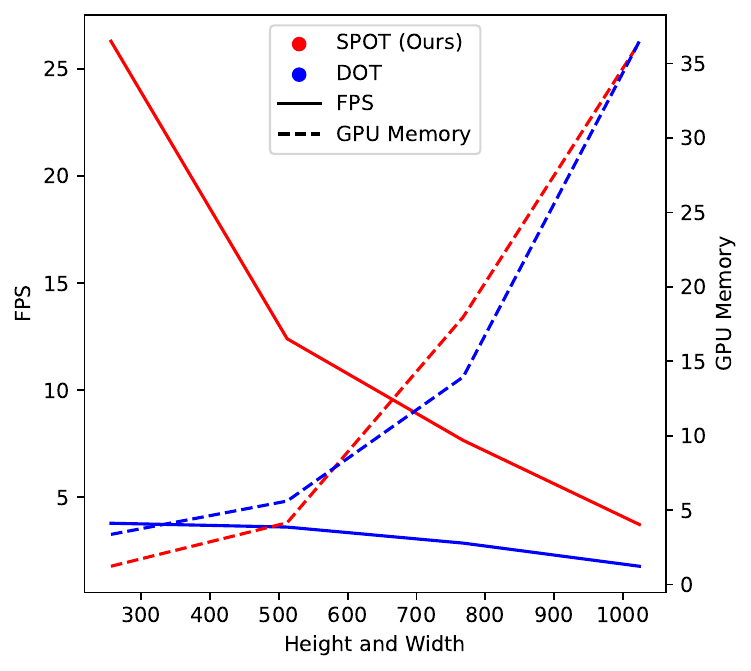}
\caption{GPU memory and inference FPS v.s. input resolution.\label{fig:complexity}}
\end{figure}

\noindent\textbf{Computational Complexity}. SPOT only maintains 3 frames memory, the computational complexity will not increase after time step $t$ being larger than 3. In addition, we provide curve of GPU memory and speed w.r.t video resolution in \cref{fig:complexity}. SPOT runs much faster than DOT across varying resolutions up to 1024x1024, while consumes similar memory. The main computational overhead still lies in RAFT. So, we believe recent progress in efficient and high-resolution optical flow estimation can greatly and immediately benefits SPOT due to the unified architecture design.

\noindent\textbf{Compared to Recent Methods}. Track-On~\cite{aydemir2025track} and DELTA~\cite{ngo2024delta} are very recent works that focus on online sparse tracking and offline dense tracking, respectively. However, SPOT tackles online dense tracking. Here, we provide an efficiency comparison with them under our setting. On H100, SPOT tracks 512x512 videos (around 262k points) at \textbf{12.4} FPS with \textbf{4.15}GB GPU memory. But Track-On can only tracks up to 21k points at \textbf{1.23} FPS with \textbf{77.2}GB  GPU memory; DELTA runs at \textbf{0.19} FPS with \textbf{49.58}GB  GPU memory. Considering the significant latency and memory consumption, both are not suitable for online dense point tracking we consider here.

\begin{figure}
\centering
\includegraphics[width=0.8\linewidth]{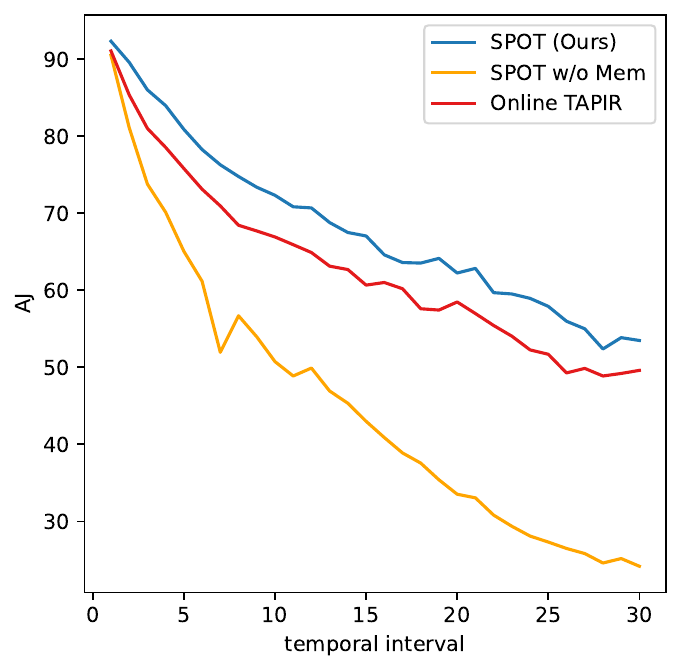}
\caption{AJ of our SPOT, SPOT w/o Memory, and Online TAPIR across varying temporal intervals on DAVIS (First).\label{fig:acc_vs_temporal_supp}}
\end{figure}

\noindent\textbf{Accuracy vs. Temporal Interval}. We repurpose TAP-Vid to evaluate performance across varying temporal intervals explicitly. Specifically, we employ the DAVIS (First) split here and evaluate the performance for each temporal interval (up to 30 here) without averaging the results across the temporal intervals. \cref{fig:acc_vs_temporal_supp} shows that AJ will degrade gradually as the temporal interval increases, while SPOT generally outperforms Online TAPIR~\cite{vecerik2024robotap} on all temporal intervals. Besides, AJ of SPOT w/o memory degrades rapidly, illustrating the important role of memory module. Though length of our memory bank is only 3, SPOT propagates information from first frame to memory frames directly, greatly alleviating the information degradation and helping handling long videos. 

\section{More Qualitative Results}

\begin{figure}
\centering
\includegraphics[width=\linewidth]{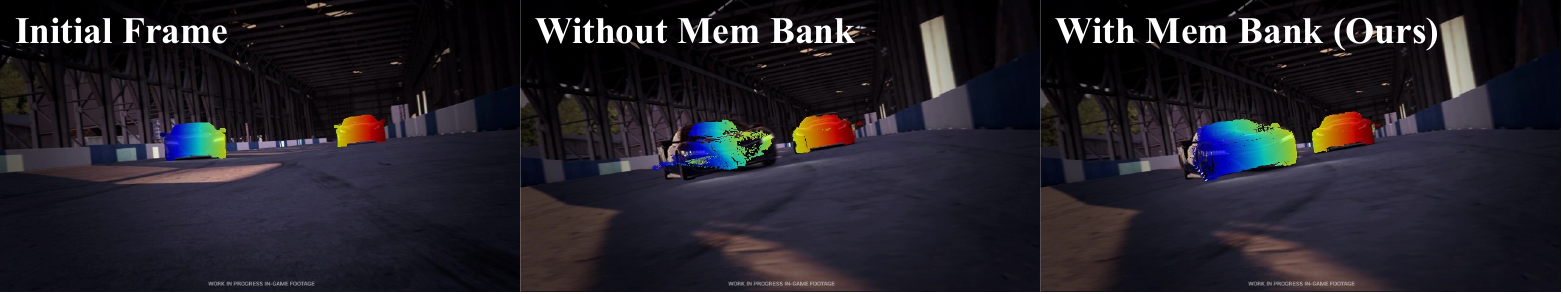}
\caption{Qualitative ablation of memory bank on challenging real-world video.\label{fig:mem_quali_supp}}
\end{figure}

\noindent\textbf{Qualitative Analysis of Memory Bank}. We provide qualitative ablation of memory bank on video with large appearance variations in \cref{fig:mem_quali_supp}. As shown in \cref{fig:mem_quali_supp}, removing the memory bank leads to the failure of tracking due to appearance changes. Our SPOT with memory bank can successfully track the points of cars and overcome the challenge of appearance variations.

\begin{figure}
\begin{centering}
\includegraphics[width=\linewidth]{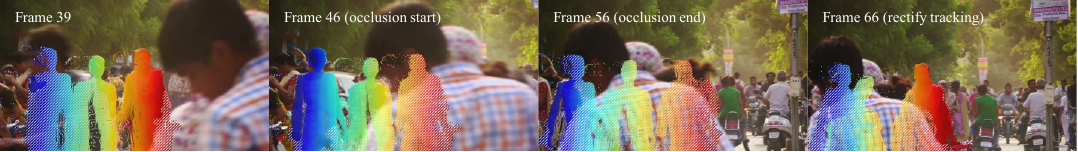}
\caption{Qualitative result on long occlusion. \label{fig:long_occ}}
\par\end{centering}
\end{figure}

\noindent\textbf{Qualitative Result on Long Occlusion}. SPOT introduces visibility mask during splatting (Eq.~7 of main paper). And splatted feature of occluded region will be all zeros and no effective information can be read out through attention. So SPOT `degrades' to pairwise method in such extreme case. We provide a such case in Fig.~\ref{fig:long_occ}, where right woman (red color) is totally occluded by man for 10 frames (longer than 3 frames memory). Once the woman reappears, SPOT successfully locates the woman and recovers from occlusions.

\begin{figure*}
\centering
\includegraphics[width=0.9\linewidth]{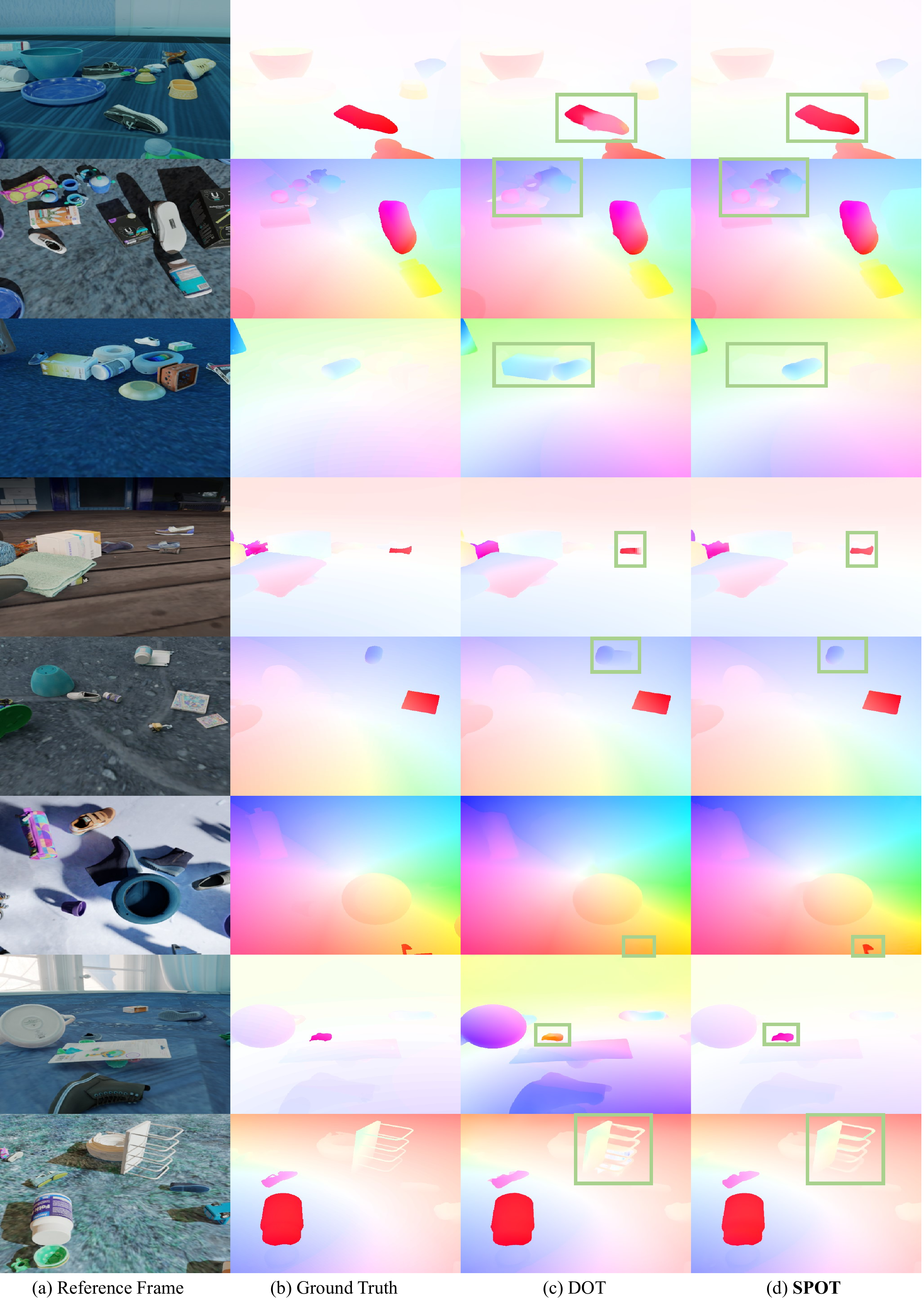}
\caption{More qualitative comparison on the CVO (Extended). Notable areas are marked by a bounding box. Please zoom in for details.\label{fig:qualitative_cvo_supp}}
\end{figure*}

\noindent\textbf{Qualitative Results on CVO}. We provide more qualitative comparison results with the previous state-of-the-art method for dense tracking, DOT~\cite{le2024dense}, on long-range optical flow benchmark CVO in \cref{fig:qualitative_cvo_supp}. The areas where our SPOT achieves substantial improvements are highlighted with bounding boxes. Please zoom in for more details. DOT typically fails to estimate the motion of small objects, occluded objects, and objects with weak textures. By contrast, our SPOT successfully estimates the motion for these hard cases.  

\begin{figure*}
\centering
\includegraphics[width=0.8\linewidth]{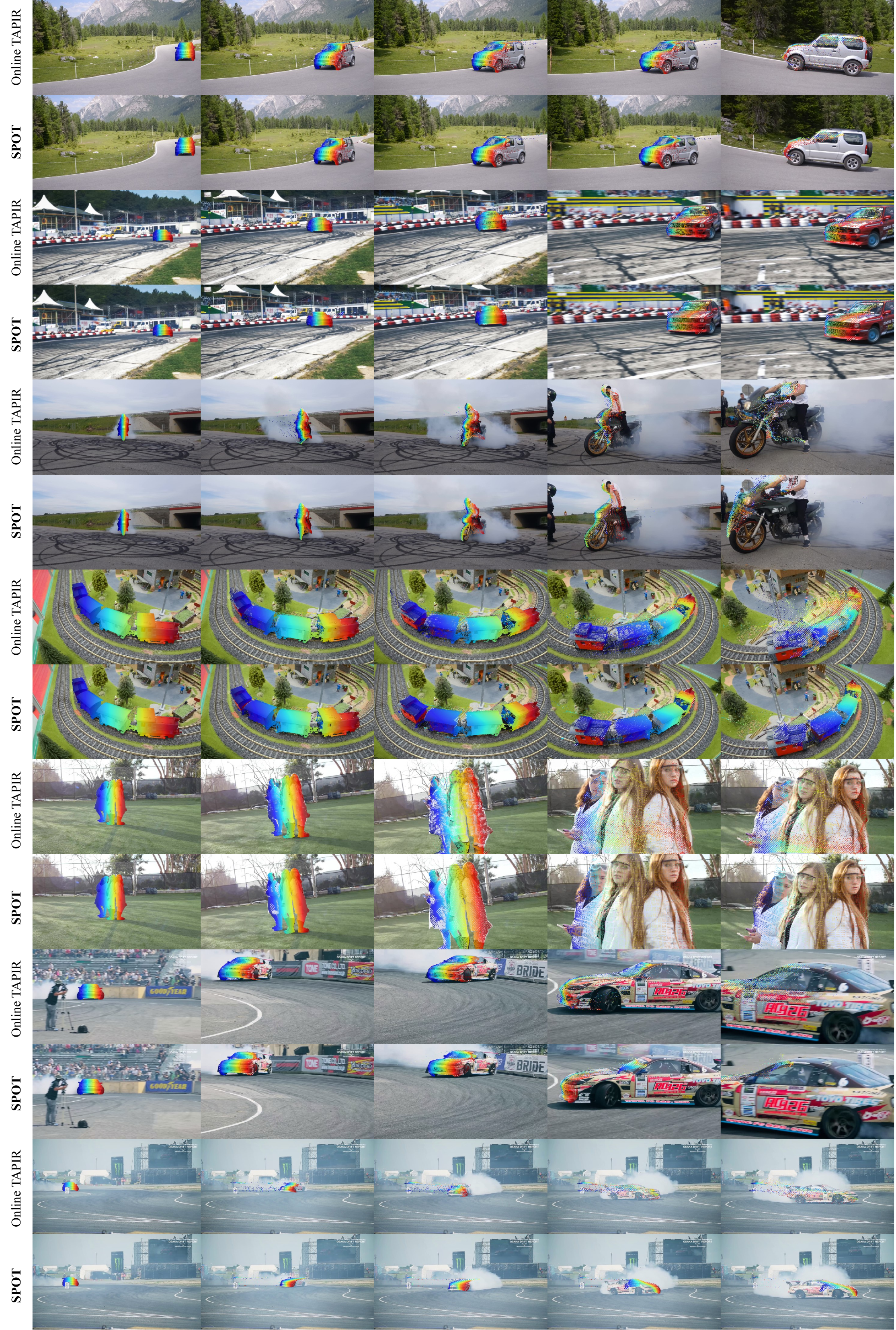}
\caption{More qualitative comparison on DAVIS. For each sequence, we show tracking results of Online TAPIR~\cite{vecerik2024robotap} and SPOT. Only foreground points of the first frame are visualized, each point is displayed with a different color and overlayed with white stripes if occluded. Please zoom in for details.\label{fig:qualitative_davis_supp}}
\end{figure*}

\noindent\textbf{Qualitative Results on Real-world Videos}. We also provide more qualitative results with the previous state-of-the-art method for online tracking, Online TAPIR~\cite{vecerik2024robotap}, on real-world videos from DAVIS in \cref{fig:qualitative_davis_supp}. \cref{fig:qualitative_davis_supp} shows that our SPOT has superior performance on real-world videos. Please zoom in for more details. 

\begin{figure*}
\centering
\includegraphics[width=0.8\linewidth]{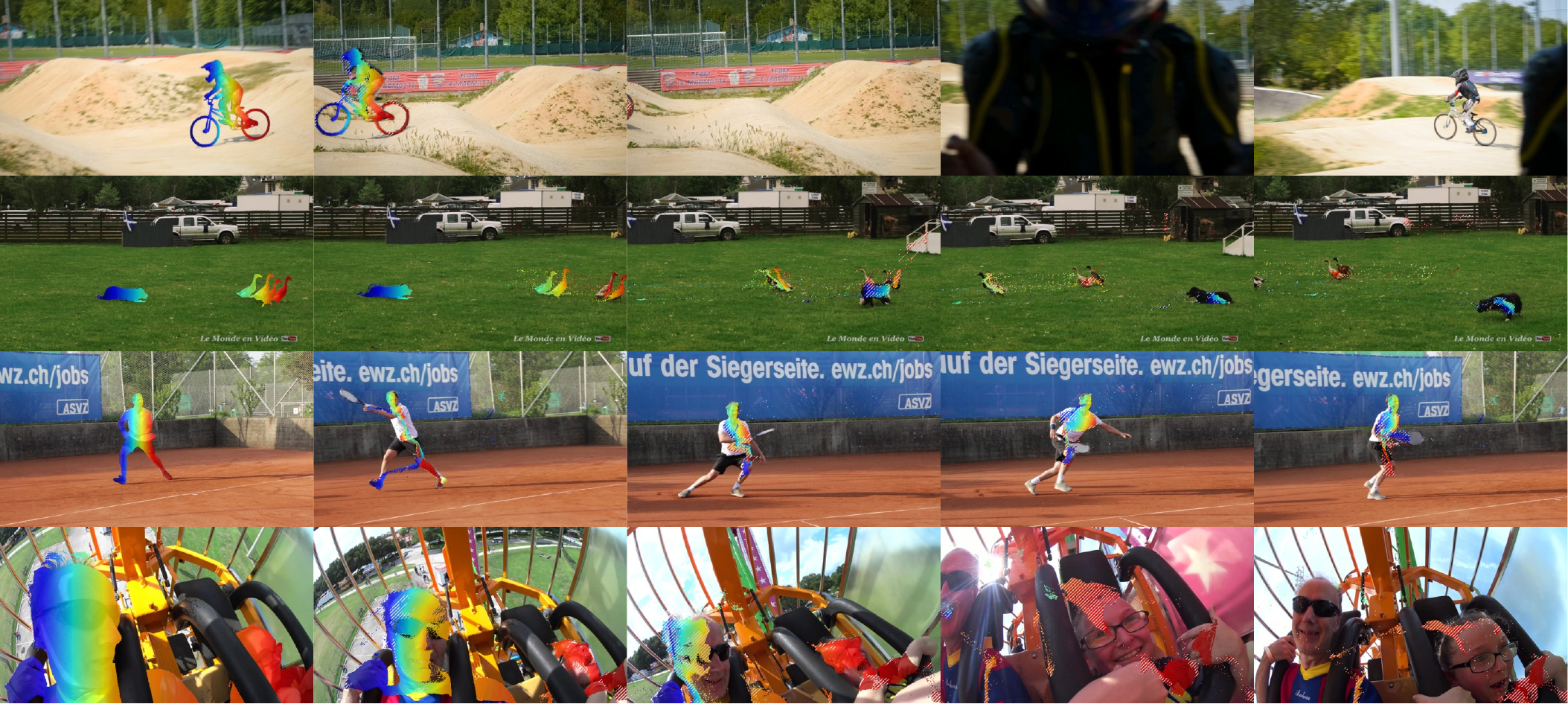}
\caption{Failure cases on DAVIS. Please zoom in for details.\label{fig:failure_case}}
\end{figure*}

\noindent\textbf{Failure Cases}. We provide some failure cases in \cref{fig:failure_case}. Our SPOT fails to track the bike after extreme long occlusion, i.e., more than 20 occluded frames in the first case. Besides, SPOT cannot distinguish texture-less objects properly, especially there are four similar fast-moving ducks in the second case. Finally, for fast motion shown in the third and fourth cases, SPOT may lose the track of thin object or even the whole object.

\end{document}